# A Parameter-efficient Convolutional Approach for Camouflaged Weed Detection in Multispectral Aerial Imagery

Leo Thomas Ramos[1,2] Angel D. Sappa[1,2,3]
[1]Computer Vision Center [2]Universitat Autònoma de Barcelona [3]ESPOL Polytechnic University
ltramos@cvc.uab.cat; sappa@ieee.org

## Abstract

*We introduce FCBNet, an efficient model designed for camouflaged weed detection. The architecture is based on a fully frozen ConvNeXt backbone, the proposed Feature Correction Block (FCB), which leverages efficient convolutions for feature refinement, and a lightweight decoder. FCBNet is evaluated on the WeedBananaCOD and WeedMap datasets under both RGB and multispectral modalities, showing that FCBNet outperforms models such as U-Net, DeepLabV3+, SK-U-Net, SegFormer, and WeedSense in terms of mIoU, exceeding 85%, while also achieving superior computational efficiency, requiring only 0.06 to 0.2 hours for training. Furthermore, the frozen backbone strategy reduces the number of trainable parameters by more than 90%, significantly lowering memory requirements. Code available at:* `https://github.com/Leo-Thomas/fcbnet`

## 1. Introduction

Weeds are unwanted camouflaged plant species that grow alongside crops and compete for essential resources such as nutrients, water, sunlight, and space [22]. As a result, weeds are recognized as one of the primary factors contributing to yield reduction across a wide range of cropping systems [5, 22], reducing overall agricultural efficiency. At a broader scale, these effects result in substantial economic losses and negatively impact productivity [28], profitability [23], and the stability of supply chains [21].

Traditional detection relies on manual field inspection [6], where farmers visually identify weed presence across cultivated areas. However, agricultural fields often extend over large areas, making this process labor-intensive [21], time-consuming [19], and difficult to perform consistently [23]. This reduces the frequency and spatial coverage of inspections, increasing the risk of delayed or incomplete detection [12]. Accordingly, this lack of precise monitoring affects subsequent management decisions [1, 34], since effective treatment depends on accurate knowledge of weed location and distribution within the field [1].

To address these limitations, Deep Learning (DL)-based approaches have gained significant attention [21, 42], particularly semantic segmentation methods [15, 21]. These models are trained on large datasets and learn discriminative patterns from visual features such as colour, texture, and shape [24]. Consequently, they enable automatic identification and precise delineation of camouflaged weeds at the pixel level [18], providing detailed information about their presence and spatial distribution. This facilitates large-scale analysis and supports more accurate and consistent monitoring compared to manual inspection [32].

Nonetheless, DL models face practical limitations. Achieving high accuracy commonly requires complex architectures with millions of trainable parameters [2], resulting in high computational and memory requirements during training [27, 28]. Also, deployment conditions impose strict constraints, since the models are often required to operate in open and remote agricultural environments [28] without access to high-performance computing infrastructure [14], and are typically integrated into platforms such as UAVs and drones with limited capacity [35]. Furthermore, multispectral/multimodal imagery is frequently employed in this domain [7, 31, 36, 37], which increases the computational requirements and data processing complexity.

Several strategies exist to obtain efficient models, but they have specific trade-offs. Model distillation transfers the knowledge from a large model to a smaller one with reduced latency [8], but it often requires a pre-trained high-capacity teacher and an distillation stage that introduces non-negligible overhead [4, 39]. Transfer learning adapts representations learned from a source domain to a target domain [20, 45], yet segmentation often requires fine-tuning a significant portion of the model [13, 46], since features for classification do not preserve the spatial and boundary information required for mask reconstruction [43, 44].

Likewise, model freezing reduces memory usage and the number of trainable parameters [40], but introduces a mismatch between fixed feature representations and the reconstruction requirements of the decoder. This often requires introducing heavy adaptation blocks to achieve competitive accuracy [11], which undermines the intended efficiency of using a frozen backbone. Therefore, developing accurate yet efficient models remains an open research challenge.

Based on the above, we propose a DL approach for camouflaged weed detection, aiming to achieve high accuracy while maintaining training efficiency and low latency. To this end, we adopt the model freezing paradigm, as it is one of the most effective methods for reducing the number of trainable parameters and computational cost. Hence, we employ a fully-frozen ConvNeXt [17] backbone integrated with a Feature Pyramid Network [16] (FPN)-based decoder for segmentation. To handle the existing mismatch, we introduce Feature Correction Blocks (FCBs), which are inserted after each ConvNeXt extraction stage to refine the features passed to the decoder. These blocks are composed of pointwise and depthwise convolutions to preserve efficiency, along with group normalization to stabilize training independently of batch size.

The proposed model is evaluated on two aerial image datasets for weed detection, covering both RGB and multispectral scenarios. The results demonstrate superior performance compared to established and state-of-the-art models. Furthermore, the use of ConvNeXt enables leveraging its stage-based design to incorporate the same number of FCBs across all its variants, regardless of their complexity. Consequently, a reduction of more than 90% in the number of trainable parameters is achieved, while maintaining superior performance and low latency. The contributions of this work are as follows:

- We propose FCBNet, an efficient model for camouflaged weed detection.
- We introduce the Feature Correction Block, which refines feature representations from a frozen backbone with minimal computational cost.
- We show that the proposed freezing strategy reduces the number of trainable parameters by more than 90%.
- We show that FCBNet achieves low latency and fast training without compromising segmentation performance.

## 2. Related work

Numerous DL approaches for weed detection have been developed in recent years. Increasing attention has been given to model performance, although not without important trade-offs. For instance, [9] proposes a semi supervised adversarial framework based on a U-Net with a ResNet50 backbone and a convolutional discriminator, enabling the use of unlabelled UAV RGB images through pseudo labelling. The method reduces annotation requirements, but the training process requires additional iterations.

The work by [35] uses a densely connected feed forward Deep Neural Network (DNN) combining handcrafted color and texture descriptors from UAV RGB images to improve weed discrimination. While it achieves high performance, the DNN training involves strenuous and prolonged processing. Similarly, [37] proposes CGS-YOLO for multispectral weed detection that incorporates CARAFE upsampling, attention mechanisms, and an additional small target detection layer to improve small instance identification. While this improves discrimination capability, parameter count and complexity increase due to the additional head and feature fusion operations.

In the same way, [31] proposes a ResNet-50-backboned U-Net combined with a fully connected Conditional Random Field (CRF) for boundary refinement. The method introduces a composite 3-channel input derived from spectral transformations to reduce computational cost. However, the CRF refinement increases post processing complexity. Likewise, [10] proposes a multi-class detection approach using U-Net and hybrid variants with pre-trained encoders (VGG19, MobileNetV2, and InceptionResNetV2). While deeper backbones achieve higher performance, they substantially increase model size and computational cost.

Other methods focus on improving efficiency to enable real-world deployment, although this often introduces performance limitations. The work by [27], for example, proposes a weed detection system based on superpixel segmentation and lightweight CNN classifiers, targeting deployment on embedded devices such as Raspberry Pi. It explores custom shallow architectures, pruning, and quantization to reduce model size and latency. However, aggressive quantization introduced accuracy degradation.

SC-Net [15] uses a U-Net based model with multi-scale convolutional blocks and attention based feature fusion to improve weed localization while reducing computational cost. Strip convolutions enlarges the receptive field with lower complexity, enabling efficient feature extraction. However, this degrades shallow feature representation and reduce performance in small regions, while the model size remains non-trivial. Similarly, [47] proposes a U-Net network with a reduced VGG backbone to improve computational efficiency. The model is pre-trained using synthetic images and later fine-tuned on real data, enabling accurate, low-latency segmentation. However, this limits feature extraction, reducing suitability for complex scenarios.

As seen, progress is driven either by strategies and components that push identification performance, or by efficiency oriented designs intended to reduce deployment cost. However, accuracy gains are often accompanied by increased training time or parameter count that raise resource demands. Conversely, approaches that prioritize compactness can reduce latency and memory footprint, but usually exhibit performance degradation under challenging conditions, highlighting the persistent trade-off between computational efficiency and robust weed identification.

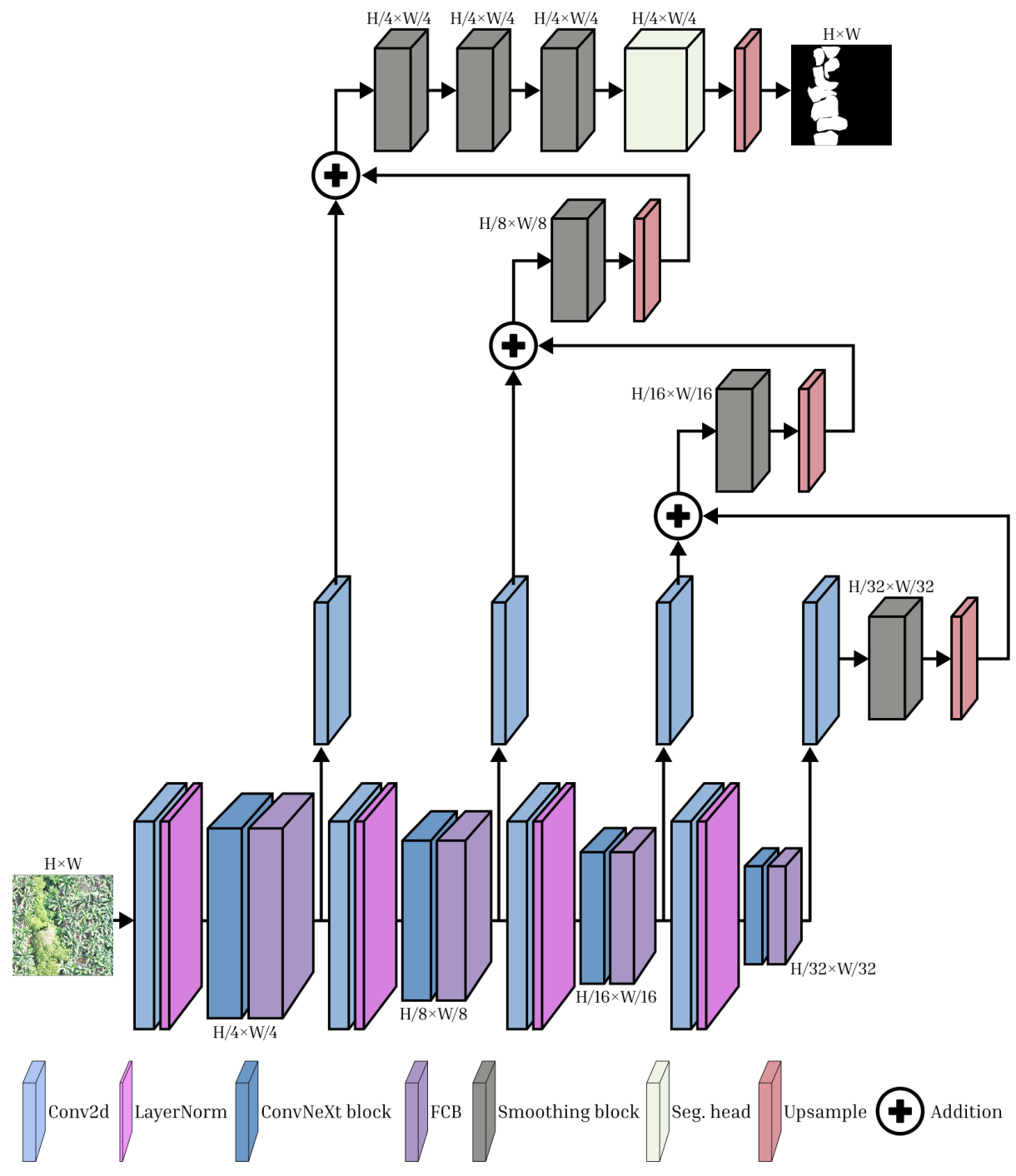

Figure 1. Overview of FCBNet proposed in this work.

# 3. Model design

The proposed model follows an encoder-decoder design, as shown in Fig. 1. It uses a fully frozen ConvNeXt backbone that produces four multi-scale feature maps, and inserts FCBs after each extraction stage to refine and correct the features. The corrected stage features are then integrated by a lightweight FPN-based decoder operating over the same four levels, which reconstructs a high-resolution representation. This representation is subsequently processed by a compact segmentation head to generate the final segmentation mask. We refer to this model as FCBNet. Below, we present each component of the architecture in detail.

## 3.1. Encoder

FCBNet's encoder is based on ConvNeXt as its underlying backbone. ConvNeXt is a Convolutional Neural Network (CNN) that builds upon ResNet50 by introducing a stage-wise design, layer normalization, GELU activations, and larger convolutional kernels (7×7). This enables effective hierarchical representation learning with controlled computational cost.

The selection of ConvNeXt in this work is motivated by its architectural design. ConvNeXt was developed with the goal of achieving performance comparable to Transformers, which currently represent the state-of-the-art in computer vision. However, Transformers exhibit quadratic complexity and require large amounts of data and training time to achieve strong performance. In response, ConvNeXt modernizes conventional CNNs through a Transformer-inspired design while preserving the efficiency advantages of CNNs.

Table 1. ConvNeXt variants and their details.

| Model | Channels (per stage) | Depths (per stage) | Params (M) |
|---|---|---|---|
| ConvNeXt-tiny | [96, 192, 384, 768] | [3, 3, 9, 3] | 28 |
| ConvNeXt-small | [96, 192, 384, 768] | [3, 3, 27, 3] | 50 |
| ConvNeXt-base | [128, 256, 512, 1024] | [3, 3, 27, 3] | 89 |
| ConvNeXt-large | [192, 384, 768, 1536] | [3, 3, 27, 3] | 198 |

Based on the above, ConvNeXt is selected as the feature extractor of FCBNet, as it provides an effective balance between the representation capability of Transformers and the computational efficiency of CNNs, which is consistent with the objective of our approach. ConvNeXt is available in four variants, as shown in Table 1. In this work, ConvNeXt-base (Fig. 2) is used as the default configuration of FCBNet.

**Feature Correction Block** As mentioned, ConvNeXt is an efficient and high-performance architecture well-suited for feature extraction. However, it has a non-trivial number of parameters, particularly in its larger variants, which can limit its suitability for deployment in resource-constrained environments. Furthermore, a segmentation model also requires the inclusion of a decoder and a segmentation head, which further increases the overall model complexity.

To maximize the efficient use of ConvNeXt, we employ it in a fully frozen configuration. This results in a substantial reduction in the number of trainable parameters, but also causes the feature representations to remain optimized for the original task and unable to be refined during training for the new domain. This, in turn, causes the decoder to operate on suboptimal feature representations, limiting its ability to accurately segment camouflaged weeds.

To address the mismatch between the fixed encoder features and the decoder requirements, we introduce the Feature Correction Block (FCB). FCB (Fig. 3) is a lightweight residual module designed to operate on convolutional feature maps. Its objective is to refine and adapt the features extracted by the backbone, enabling more effective integration by the decoder while preserving computational efficiency.

A FCB comprises seven layers in a bottleneck-like structure: a Pointwise convolution (PWConv), followed by

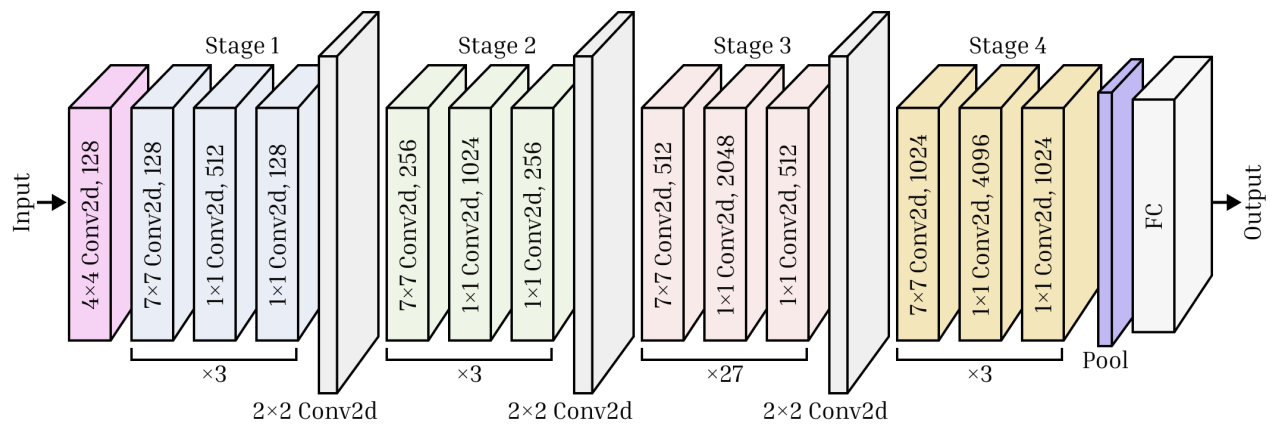

Figure 2. ConvNeXt-base architecture. Adapted from: [26]

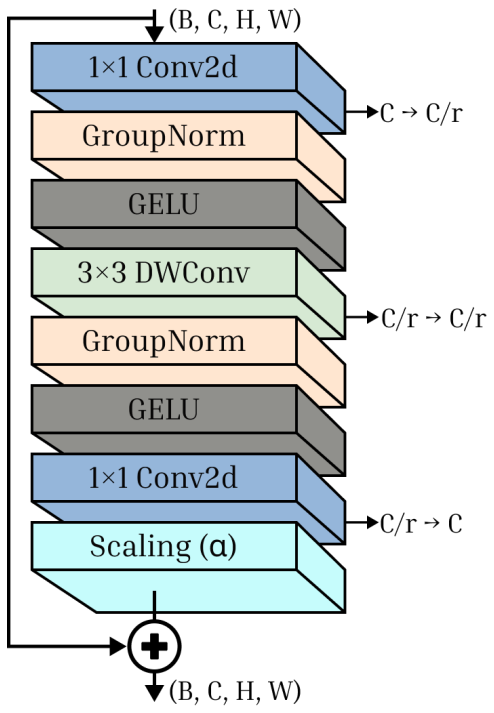

Figure 3. Structure of the Feature Correction Block.

Group Normalization (GroupNorm) with GELU activation, a Depthwise convolution (DWConv), a second GroupNorm + GELU, and a final PWConv. In addition, each FCB incorporates a residual connection that links the input and output of each block, that allows it to introduce corrective modifications while preserving the original feature information.

PWConv and DWConv are a design choice for computational efficiency. PWConv ($1\times1$) enables channel-wise feature projection with minimal overhead, while DWConv captures spatial context at significantly lower cost than standard convolutions. Likewise, GroupNorm is used to stabilize optimization without relying on large batch sizes, which is particularly useful under resource constrained training settings.

Mathematically, FCB implements a transformation $f(\cdot)$ over the input tensor $x \in \mathbb{R}^{B\times C\times H\times W}$, as in Eq. 1:

$$y = x + \alpha f(x), \tag{1}$$

where $f(\cdot)$ operates in a lower dimensional embedded space, applies spatial processing, and produces a correction term that is added to the input through a residual connection scaled by the learnable parameter $\alpha \in \mathbb{R}$.

The transformation $f(x)$ follows a bottleneck structure composed of efficient convolutional operations. First, the input tensor with $C$ channels is projected into an intermediate space of dimensionality $C'$ defined in Eq. 2:

$$C' = \max\left(C_{\min}, \left\lfloor \frac{C}{r} \right\rfloor\right), \tag{2}$$

where $r$ corresponds to the bottleneck ratio controlling the channel reduction, and $C_{\min}$ is a lower bound imposed to ensure sufficient representational capacity. This projection is implemented using a bias free $1\times1$ convolution, followed by GroupNorm and a GELU activation, as in Eq. 3:

$$z_1 = \phi\left(\text{GroupNorm}_1\left(\text{Conv}_{1\times1}^{C\to C'}(x)\right)\right), \tag{3}$$

where, $\phi(\cdot)$ denotes the GELU activation function. Next, a DWConv with kernel size $k \times k$ is applied to model spatial context efficiently, again followed by GroupNorm and GELU, as shown in Eq. 4:

$$z_2 = \phi\Big(\text{GroupNorm}_2\Big(\text{DWConv}_{k\times k}(z_1)\Big)\Big). \tag{4}$$

Finally, the representation is projected back to the original channel dimensionality using another $1\times1$ convolution, as in Eq. 5:

$$f(x) = \text{Conv}_{1\times1}^{C'\to C}(z_2). \tag{5}$$

In our implementation, the number of GroupNorm groups is dynamically selected to exactly divide the number of channels $C'$, preventing invalid configurations when $C'$ is small or not divisible by the preferred number of groups. The entire refining process is achieved using lightweight operations, avoiding the heavy parameter and computational overhead of alternative mechanisms like attention modules.

Within the frozen ConvNeXt encoder, one FCB is inserted immediately after each feature extraction stage, allowing progressive correction of the multi-scale representations. Since all ConvNeXt variants share the same four-stage hierarchical structure, the number of FCBs remains constant regardless of the backbone complexity.

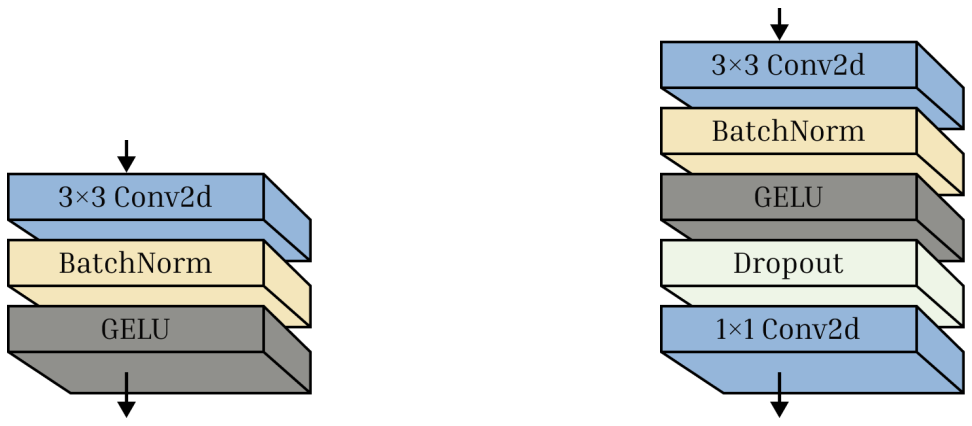

(a) Smoothing block. (b) Segmentation head.

Figure 4. Smoothing block and head used in FCBNet decoder.

### 3.2. Decoder and head

To fuse the features extracted by the encoder, we employ a lightweight FPN-based decoder. It integrates four multi-scale feature maps into a single, high-resolution representation optimized for segmentation. Additionally, the design is streamlined to maintain computational efficiency.

Specifically, it includes four $1\times1$ convolutions that project the feature maps from each encoder stage into a shared channel dimensionality. These are followed by four smoothing blocks (Fig. 4a), each consisting of a $3\times3$ convolution, BatchNorm, and GELU activation. These blocks are used during the top-down fusion process, where lower resolution features are upsampled using bilinear interpolation and combined with higher resolution features through element wise summation. Two additional smoothing blocks form a refinement stage applied at the highest resolution to further consolidate the decoded features before prediction.

Finally, a compact segmentation head (Fig. 4b) converts the decoder features into class predictions. The head uses a 3×3 convolution followed by BatchNorm[1] and GELU, applies dropout for regularization, and then uses a final 1×1 convolution to map features to the desired number of output channels. The resulting logits are upsampled to the original input resolution using bilinear interpolation, producing the final full-resolution segmentation mask.

# 4. Experiments

## 4.1. Datasets and preprocessing

To evaluate the proposed model, two aerial imagery datasets, WeedBananaCOD [38] and WeedMap [30], for weed detection are selected.

WeedBananaCOD [38] is an aerial dataset for camouflaged weed detection. The images were acquired over an approximately 4.3 hectare banana field located in the coastal region of Ecuador. The site is characterized by dense and homogeneous vegetation, creating challenging conditions where weeds are often visually camouflaged within the surrounding crop. Images were captured using a MicaSense RedEdge-M camera mounted on a DJI Mavic 2 Pro UAV. The dataset is multispectral, including RGB and near infrared (NIR) bands, and comprises a tile of size 5571×5855 pixels. The images were divided into 512×512 patches using the official preprocessing code provided by the authors.

WeedMap [30] is one of the best-known datasets for weed detection. It was acquired in sugar beet fields located in Switzerland and Germany. It consists of two subsets: one captured with a RedEdge-M camera mounted on a DJI Inspire 2 UAV, providing five spectral bands (RGB, NIR, and red edge (RE)), and another by a Sequoia camera mounted on a DJI Mavic Pro UAV with RGB-NIR configuration. In this work, the RedEdge-M subset[2] (5964 images of 480×360 pixels) was selected, as it contains a larger number of images with weed presence, while the Sequoia subset includes a higher proportion of other vegetation classes.

## 4.2. Experimental setup

Training is conducted in a Python environment using cross entropy loss and the AdamW optimizer. The learning rate is set to 1e-3 and scheduled using a OneCycle policy. Models were trained for 100 epochs with a batch size of 64 on an NVIDIA A100 GPU with 40 GB of memory. For comparison, several semantic segmentation models were trained under the same configuration, including established architectures and weed specific methods such as U-Net [29], DeepLabV3+ [3], SegFormer [41], SK-UNet [25], and WeedSense [33].

Performance is evaluated using the main semantic segmentation metrics, including intersection over union (IoU) for weed and background classes, and mean IoU (mIoU). Computational efficiency is assessed by measuring inference time (average latency per image), training time, total number of parameters, number of trainable parameters, and GFLOPs. Additionally, precision, recall, and F1-score were computed as part of the full evaluation, with detailed results provided in the supplementary material. The experiments were conducted using both RGB and full spectral configurations. To support additional channels, the first layer of each model is expanded accordingly.

Table 2. Ablation of the FCB $\alpha$ parameter for feature refinement.

| Alpha | Dataset | Modality | $IoU_{BG}\uparrow$ | $IoU_{W}\uparrow$ | mIoU↑ | Inference time (s) | Training time (h) | Total params. (M) | Trainable params. (M) | GFLOPS |
|---|---|---|---|---|---|---|---|---|---|---|
| 0.01 | WeedBananaCOD | RGB | 0.990 | 0.730 | 0.860 | 0.0236 | 0.101 | 91.277 | 2.685 | 15.156 |
| | | RGB-NIR | 0.991 | 0.754 | 0.873 | 0.0237 | 0.106 | 91.279 | 2.685 | 15.190 |
| | WeedMap | RGB | 0.985 | 0.531 | 0.758 | 0.0141 | 0.129 | 91.277 | 2.685 | 10.657 |
| | | RGB-NIR | 0.986 | 0.531 | 0.759 | 0.0144 | 0.137 | 91.279 | 2.685 | 10.680 |
| | | RGB-NIR-RE | 0.986 | 0.536 | 0.761 | 0.0147 | 0.144 | 91.281 | 2.685 | 10.704 |
| 0.03 | WeedBananaCOD | RGB | 0.990 | 0.732 | 0.861 | 0.0253 | 0.103 | 91.277 | 2.685 | 15.156 |
| | | RGB-NIR | 0.991 | 0.747 | 0.869 | 0.0279 | 0.108 | 91.279 | 2.685 | 15.190 |
| | WeedMap | RGB | 0.983 | 0.529 | 0.756 | 0.0173 | 0.132 | 91.277 | 2.685 | 10.657 |
| | | RGB-NIR | 0.986 | 0.531 | 0.759 | 0.0175 | 0.140 | 91.279 | 2.685 | 10.680 |
| | | RGB-NIR-RE | 0.986 | 0.535 | 0.761 | 0.0179 | 0.146 | 91.281 | 2.685 | 10.704 |
| 0.05 | WeedBananaCOD | RGB | 0.990 | 0.722 | 0.856 | 0.0254 | 0.102 | 91.277 | 2.685 | 15.156 |
| | | RGB-NIR | 0.992 | 0.760 | 0.876 | 0.0281 | 0.109 | 91.279 | 2.685 | 15.190 |
| | WeedMap | RGB | 0.985 | 0.527 | 0.756 | 0.0158 | 0.130 | 91.277 | 2.685 | 10.657 |
| | | RGB-NIR | 0.985 | 0.530 | 0.758 | 0.0164 | 0.138 | 91.279 | 2.685 | 10.680 |
| | | RGB-NIR-RE | 0.985 | 0.532 | 0.759 | 0.0169 | 0.145 | 91.281 | 2.685 | 10.704 |
| **0.07** | WeedBananaCOD | RGB | 0.991 | 0.733 | 0.862 | 0.0227 | 0.099 | 91.277 | 2.685 | 15.156 |
| | | RGB-NIR | 0.991 | 0.762 | 0.877 | 0.0298 | 0.105 | 91.279 | 2.685 | 15.190 |
| | WeedMap | RGB | 0.985 | 0.534 | 0.760 | 0.0141 | 0.128 | 91.277 | 2.685 | 10.657 |
| | | RGB-NIR | 0.986 | 0.545 | 0.766 | 0.0145 | 0.134 | 91.279 | 2.685 | 10.680 |
| | | RGB-NIR-RE | 0.987 | 0.547 | 0.767 | 0.0147 | 0.142 | 91.281 | 2.685 | 10.704 |
| 0.09 | WeedBananaCOD | RGB | 0.990 | 0.713 | 0.851 | 0.0215 | 0.104 | 91.277 | 2.685 | 15.156 |
| | | RGB-NIR | 0.991 | 0.749 | 0.870 | 0.0240 | 0.105 | 91.279 | 2.685 | 15.190 |
| | WeedMap | RGB | 0.984 | 0.532 | 0.758 | 0.0169 | 0.130 | 91.277 | 2.685 | 10.657 |
| | | RGB-NIR | 0.985 | 0.531 | 0.758 | 0.0177 | 0.143 | 91.279 | 2.685 | 10.680 |
| | | RGB-NIR-RE | 0.986 | 0.537 | 0.762 | 0.0172 | 0.144 | 91.281 | 2.685 | 10.704 |

## 4.3. Results and discussion

Table 2 evaluates the impact of the $\alpha$ parameter on model performance. Results show a modest but consistent improvement as $\alpha$ increases, with the best performance achieved at $\alpha$ = 0.07 across both RGB and multispectral configurations. This indicates that stronger feature correction improves representation quality. Beyond 0.07, performance degrades, suggesting that excessive correction disrupts feature stability rather than enhancing discrimination. Also, $\alpha$ has no effect on efficiency, as parameter count, GFLOPs, and latency remain constant.

The ablation of the FCB bottleneck ratio is shown in Table 3. The results indicate that a configuration of 2 represents the optimal balance point. Due to the inverse nature of this parameter in the FCB block, increasing this ratio reduces model complexity and accelerates computation, as higher values decrease the number of internal channels. However, ratios above 2 lead to performance degradation. The excessive reduction in internal dimensionality limits the model's representational capacity, preventing the block from capturing the fine details required for segmenta-tion and resulting in a gradual loss of accuracy.

[1]By design, FCBNet employs a hybrid normalization. GroupNorm is used in the frozen backbone to avoid using mismatched pre-trained batch statistics, while BatchNorm is used in the trainable decoder to leverage mini-batch information for faster convergence and optimize gradient flow.

[2]Note that WeedMap contains multiple semantic classes; for this study, all non-weed classes were merged into the background to enable binary segmentation.

Table 3. Ablation of the of the FCB bottleneck ratio.

| Bottleneck ratio | Dataset | Modality | IoU$_{BG}$ ↑ | IoU$_{W}$ ↑ | mIoU↑ | Inference time (s) | Training time (h) | Total params. (M) | Trainable params. (M) | GFLOPS |
|---|---|---|---|---|---|---|---|---|---|---|
| 1 | WeedBananaCOD | RGB | 0.991 | 0.739 | 0.865 | 0.0248 | 0.107 | 92.682 | 4.090 | 16.260 |
| | | RGB-NIR | 0.991 | 0.752 | 0.872 | 0.0306 | 0.112 | 92.684 | 4.090 | 16.293 |
| | WeedMap | RGB | 0.985 | 0.517 | 0.751 | 0.0167 | 0.135 | 92.682 | 4.090 | 11.432 |
| | | RGB-NIR | 0.984 | 0.520 | 0.752 | 0.0169 | 0.139 | 92.684 | 4.090 | 11.456 |
| | | RGB-NIR-RE | 0.985 | 0.537 | 0.761 | 0.0171 | 0.148 | 92.686 | 4.090 | 11.480 |
| **2** | WeedBananaCOD | RGB | 0.991 | 0.733 | 0.862 | 0.0227 | 0.099 | 91.277 | 2.685 | 15.156 |
| | | RGB-NIR | 0.991 | 0.762 | 0.877 | 0.0298 | 0.105 | 91.279 | 2.685 | 15.190 |
| | WeedMap | RGB | 0.985 | 0.534 | 0.760 | 0.0141 | 0.128 | 91.277 | 2.685 | 10.657 |
| | | RGB-NIR | 0.986 | 0.545 | 0.766 | 0.0145 | 0.134 | 91.279 | 2.685 | 10.680 |
| | | RGB-NIR-RE | 0.987 | 0.547 | 0.767 | 0.0147 | 0.142 | 91.281 | 2.685 | 10.704 |
| 3 | WeedBananaCOD | RGB | 0.988 | 0.728 | 0.858 | 0.0220 | 0.096 | 90.812 | 2.221 | 14.882 |
| | | RGB-NIR | 0.990 | 0.753 | 0.872 | 0.0276 | 0.102 | 90.814 | 2.221 | 14.915 |
| | WeedMap | RGB | 0.985 | 0.526 | 0.756 | 0.0137 | 0.121 | 90.812 | 2.221 | 10.464 |
| | | RGB-NIR | 0.985 | 0.538 | 0.762 | 0.0138 | 0.126 | 90.814 | 2.221 | 10.487 |
| | | RGB-NIR-RE | 0.986 | 0.543 | 0.764 | 0.0140 | 0.129 | 90.816 | 2.221 | 10.511 |
| 4 | WeedBananaCOD | RGB | 0.990 | 0.729 | 0.859 | 0.0214 | 0.094 | 90.583 | 1.991 | 14.747 |
| | | RGB-NIR | 0.991 | 0.740 | 0.865 | 0.0258 | 0.097 | 90.585 | 1.991 | 14.780 |
| | WeedMap | RGB | 0.985 | 0.513 | 0.749 | 0.0127 | 0.118 | 90.583 | 1.991 | 10.369 |
| | | RGB-NIR | 0.985 | 0.527 | 0.756 | 0.0127 | 0.123 | 90.585 | 1.991 | 10.393 |
| | | RGB-NIR-RE | 0.985 | 0.530 | 0.758 | 0.0129 | 0.124 | 90.587 | 1.991 | 10.416 |
| 5 | WeedBananaCOD | RGB | 0.989 | 0.688 | 0.838 | 0.0214 | 0.090 | 90.449 | 1.857 | 14.692 |
| | | RGB-NIR | 0.989 | 0.702 | 0.846 | 0.0248 | 0.095 | 90.451 | 1.857 | 14.725 |
| | WeedMap | RGB | 0.985 | 0.517 | 0.751 | 0.0125 | 0.114 | 90.449 | 1.857 | 10.330 |
| | | RGB-NIR | 0.985 | 0.525 | 0.755 | 0.0123 | 0.125 | 90.451 | 1.857 | 10.354 |
| | | RGB-NIR-RE | 0.986 | 0.530 | 0.758 | 0.0128 | 0.124 | 90.453 | 1.857 | 10.377 |

The ablation study on the kernel size (Table 4) of the DWConvs in the FCB shows that a 3×3 configuration achieves the best performance. Although larger kernels (5×5 and 7×7) expand the receptive field and capture broader context, this becomes counterproductive within the FCB block. Since the backbone (ConvNeXt) already extracts large scale contextual information (7×7 kernels), using larger kernels in the FCB appears to introduce redundancy that degrades local detail sharpness. Additionally, increasing the kernel size leads to higher latency, training time, and GFLOPs without improving performance.

The ablation study on the FPN feature dimension (Table 5) shows that a dimension of 128 represents the optimal balance between performance and efficiency. A non-linear behavior is observed, where performance drops when increasing to 160, followed by a slight recovery at 192 without reaching the previous peak. This phenomenon suggests that increasing the dimensionality beyond 128 introduces structural redundancy that initially hinders convergence and dilutes encoder signals. Although a capacity of 192 allows the model to reorganize part of this information, the exponential increase in GFLOPs (from 15.1 to 28.4 for WeedBananaCOD) and latency confirms that higher dimensions saturate the decoder without providing meaningful improvements in data representational quality.

Table 4. Ablation of the depthwise kernel size in the FCB block.

| Kernel size | Dataset | Modality | IoU$_{BG}$ ↑ | IoU$_{W}$ ↑ | mIoU↑ | Inference time (s) | Training time (h) | Total params. (M) | Trainable params. (M) | GFLOPS |
|---|---|---|---|---|---|---|---|---|---|---|
| 1 | WeedBananaCOD | RGB | 0.989 | 0.696 | 0.842 | 0.0225 | 0.096 | 91.269 | 2.677 | 15.141 |
| | | RGB-NIR | 0.988 | 0.677 | 0.833 | 0.0296 | 0.103 | 91.271 | 2.677 | 15.174 |
| | WeedMap | RGB | 0.986 | 0.535 | 0.760 | 0.0130 | 0.123 | 91.269 | 2.677 | 10.646 |
| | | RGB-NIR | 0.985 | 0.533 | 0.759 | 0.0137 | 0.135 | 91.271 | 2.677 | 10.669 |
| | | RGB-NIR-RE | 0.986 | 0.537 | 0.761 | 0.0139 | 0.140 | 91.273 | 2.677 | 10.693 |
| **3** | WeedBananaCOD | RGB | 0.991 | 0.733 | 0.862 | 0.0227 | 0.099 | 91.277 | 2.685 | 15.156 |
| | | RGB-NIR | 0.991 | 0.762 | 0.877 | 0.0298 | 0.105 | 91.279 | 2.685 | 15.190 |
| | WeedMap | RGB | 0.985 | 0.534 | 0.760 | 0.0141 | 0.128 | 91.277 | 2.685 | 10.657 |
| | | RGB-NIR | 0.986 | 0.545 | 0.766 | 0.0145 | 0.134 | 91.279 | 2.685 | 10.680 |
| | | RGB-NIR-RE | 0.987 | 0.547 | 0.767 | 0.0147 | 0.142 | 91.281 | 2.685 | 10.704 |
| 5 | WeedBananaCOD | RGB | 0.990 | 0.724 | 0.857 | 0.0248 | 0.103 | 91.292 | 2.700 | 15.188 |
| | | RGB-NIR | 0.990 | 0.753 | 0.872 | 0.0300 | 0.107 | 91.294 | 2.700 | 15.221 |
| | WeedMap | RGB | 0.986 | 0.532 | 0.759 | 0.0168 | 0.132 | 91.292 | 2.700 | 10.679 |
| | | RGB-NIR | 0.986 | 0.540 | 0.763 | 0.0169 | 0.141 | 91.294 | 2.700 | 10.703 |
| | | RGB-NIR-RE | 0.986 | 0.543 | 0.765 | 0.0173 | 0.147 | 91.269 | 2.700 | 10.726 |
| 7 | WeedBananaCOD | RGB | 0.988 | 0.682 | 0.835 | 0.0253 | 0.105 | 91.315 | 2.724 | 15.235 |
| | | RGB-NIR | 0.991 | 0.736 | 0.863 | 0.0309 | 0.109 | 91.317 | 2.724 | 15.269 |
| | WeedMap | RGB | 0.985 | 0.527 | 0.756 | 0.0176 | 0.132 | 91.315 | 2.724 | 10.712 |
| | | RGB-NIR | 0.986 | 0.535 | 0.761 | 0.0178 | 0.144 | 91.317 | 2.724 | 10.736 |
| | | RGB-NIR-RE | 0.986 | 0.533 | 0.760 | 0.0174 | 0.147 | 91.319 | 2.724 | 10.759 |

Table 5. Ablation study of the FPN feature dimension.

| Feature dim. | Dataset | Modality | IoU$_{BG}$ ↑ | IoU$_{W}$ ↑ | mIoU↑ | Inference time (s) | Training time (h) | Total params. (M) | Trainable params. (M) | GFLOPS |
|---|---|---|---|---|---|---|---|---|---|---|
| 96 | WeedBananaCOD | RGB | 0.990 | 0.717 | 0.853 | 0.0222 | 0.093 | 90.763 | 2.172 | 10.448 |
| | | RGB-NIR | 0.991 | 0.739 | 0.865 | 0.0292 | 0.095 | 90.765 | 2.172 | 10.482 |
| | WeedMap | RGB | 0.986 | 0.524 | 0.755 | 0.0130 | 0.126 | 90.763 | 2.172 | 7.346 |
| | | RGB-NIR | 0.986 | 0.530 | 0.758 | 0.0132 | 0.133 | 90.765 | 2.172 | 7.370 |
| | | RGB-NIR-RE | 0.984 | 0.537 | 0.761 | 0.0135 | 0.142 | 90.767 | 2.172 | 7.393 |
| **128** | WeedBananaCOD | RGB | 0.991 | 0.733 | 0.862 | 0.0227 | 0.099 | 91.277 | 2.685 | 15.156 |
| | | RGB-NIR | 0.991 | 0.762 | 0.877 | 0.0298 | 0.105 | 91.279 | 2.685 | 15.190 |
| | WeedMap | RGB | 0.985 | 0.534 | 0.760 | 0.0141 | 0.128 | 91.277 | 2.685 | 10.657 |
| | | RGB-NIR | 0.986 | 0.545 | 0.766 | 0.0145 | 0.134 | 91.279 | 2.685 | 10.680 |
| | | RGB-NIR-RE | 0.987 | 0.547 | 0.767 | 0.0147 | 0.142 | 91.281 | 2.685 | 10.704 |
| 160 | WeedBananaCOD | RGB | 0.989 | 0.719 | 0.854 | 0.0252 | 0.105 | 91.919 | 3.328 | 21.172 |
| | | RGB-NIR | 0.990 | 0.748 | 0.869 | 0.0308 | 0.108 | 91.921 | 3.328 | 21.205 |
| | WeedMap | RGB | 0.986 | 0.526 | 0.756 | 0.0170 | 0.132 | 91.919 | 3.328 | 14.886 |
| | | RGB-NIR | 0.983 | 0.535 | 0.759 | 0.0171 | 0.143 | 91.920 | 3.328 | 14.910 |
| | | RGB-NIR-RE | 0.986 | 0.541 | 0.764 | 0.0177 | 0.151 | 91.923 | 3.328 | 14.934 |
| 192 | WeedBananaCOD | RGB | 0.990 | 0.731 | 0.861 | 0.0260 | 0.107 | 92.691 | 4.099 | 28.494 |
| | | RGB-NIR | 0.991 | 0.750 | 0.871 | 0.0316 | 0.110 | 92.693 | 4.099 | 28.528 |
| | WeedMap | RGB | 0.983 | 0.527 | 0.755 | 0.0170 | 0.131 | 92.691 | 4.099 | 20.035 |
| | | RGB-NIR | 0.986 | 0.542 | 0.764 | 0.0173 | 0.147 | 92.693 | 4.099 | 20.059 |
| | | RGB-NIR-RE | 0.985 | 0.545 | 0.765 | 0.0179 | 0.153 | 92.695 | 4.099 | 20.082 |

The ablation of refine depth in Table 6 shows that integrating two refinement blocks achieves the optimal balance between noise suppression in the fused FPN features and preservation of spatial detail. Exceeding this threshold leads to performance degradation. This is likely due to excessive smoothing blocks introducing an over-smoothing effect that blurs fine details, saturating the model's capacity and leading to overfitting without improving representational quality, despite the increase in parameters and the associated rise in computational cost.

The ablation shown in Table 7 demonstrates that integrating the FCB block is the determining factor for achieving optimal performance. The results indicate that the w/FCB configuration achieves the highest mIoU across all datasets and spectral modalities, significantly outperform-ing both the baseline model (w/o FCB) and the alternative using CBAM. When the FCB is removed, a marked performance drop is observed, indicating that the frozen backbone alone lacks the necessary adaptation capacity for multispectral segmentation. In contrast, replacing the FCB with the CBAM attention block does not provide meaningful benefits; although CBAM presents fewer theoretical parameters and GFLOPs, this does not translate into real efficiency, resulting in noticeably higher training and inference times. This behavior confirms that CBAM operations are more costly in practice, while the FCB achieves the best balance between robustness, lower latency, and prediction accuracy.

Table 6. Ablation of the decoder refinement depth.

| Refine depth | Dataset | Modality | IoU$_{BG}$ ↑ | IoU$_{W}$ ↑ | mIoU↑ | Inference time (s) | Training time (h) | Total params. (M) | Trainable params. (M) | GFLOPS |
|---|---|---|---|---|---|---|---|---|---|---|
| 0 | WeedBananaCOD | RGB | 0.990 | 0.716 | 0.853 | 0.0219 | 0.093 | 90.981 | 2.390 | 10.312 |
| | | RGB-NIR | 0.990 | 0.731 | 0.860 | 0.0226 | 0.094 | 90.983 | 2.390 | 10.346 |
| | WeedMap | RGB | 0.985 | 0.509 | 0.747 | 0.0140 | 0.131 | 90.981 | 2.390 | 7.251 |
| | | RGB-NIR | 0.986 | 0.521 | 0.753 | 0.0144 | 0.135 | 90.983 | 2.390 | 7.274 |
| | | RGB-NIR-RE | 0.986 | 0.530 | 0.758 | 0.0146 | 0.135 | 90.985 | 2.390 | 7.298 |
| 1 | WeedBananaCOD | RGB | 0.990 | 0.713 | 0.852 | 0.0219 | 0.095 | 91.129 | 2.537 | 12.734 |
| | | RGB-NIR | 0.991 | 0.741 | 0.866 | 0.0229 | 0.097 | 91.131 | 2.537 | 12.768 |
| | WeedMap | RGB | 0.984 | 0.520 | 0.752 | 0.0145 | 0.133 | 91.129 | 2.537 | 8.954 |
| | | RGB-NIR | 0.985 | 0.527 | 0.756 | 0.0146 | 0.133 | 91.131 | 2.537 | 8.977 |
| | | RGB-NIR-RE | 0.986 | 0.539 | 0.762 | 0.0147 | 0.139 | 91.133 | 2.537 | 9.001 |
| **2** | WeedBananaCOD | RGB | 0.991 | 0.733 | 0.862 | 0.0227 | 0.099 | 91.277 | 2.685 | 15.156 |
| | | RGB-NIR | 0.991 | 0.762 | 0.877 | 0.0298 | 0.105 | 91.279 | 2.685 | 15.190 |
| | WeedMap | RGB | 0.985 | 0.534 | 0.760 | 0.0141 | 0.128 | 91.277 | 2.685 | 10.657 |
| | | RGB-NIR | 0.986 | 0.545 | 0.766 | 0.0145 | 0.134 | 91.279 | 2.685 | 10.680 |
| | | RGB-NIR-RE | 0.987 | 0.547 | 0.767 | 0.0147 | 0.142 | 91.281 | 2.685 | 10.704 |
| 3 | WeedBananaCOD | RGB | 0.990 | 0.725 | 0.857 | 0.0251 | 0.105 | 91.424 | 2.833 | 17.579 |
| | | RGB-NIR | 0.991 | 0.753 | 0.872 | 0.0278 | 0.107 | 91.426 | 2.833 | 17.612 |
| | WeedMap | RGB | 0.984 | 0.504 | 0.744 | 0.0150 | 0.131 | 91.424 | 2.833 | 12.360 |
| | | RGB-NIR | 0.986 | 0.540 | 0.763 | 0.0154 | 0.141 | 91.426 | 2.833 | 12.383 |
| | | RGB-NIR-RE | 0.986 | 0.543 | 0.764 | 0.0157 | 0.145 | 91.428 | 2.833 | 12.407 |
| 4 | WeedBananaCOD | RGB | 0.989 | 0.693 | 0.841 | 0.0249 | 0.106 | 91.572 | 2.981 | 20.001 |
| | | RGB-NIR | 0.990 | 0.737 | 0.864 | 0.0263 | 0.107 | 91.574 | 2.981 | 20.034 |
| | WeedMap | RGB | 0.985 | 0.519 | 0.752 | 0.0153 | 0.138 | 91.572 | 2.981 | 14.063 |
| | | RGB-NIR | 0.986 | 0.531 | 0.759 | 0.0153 | 0.142 | 91.574 | 2.981 | 14.087 |
| | | RGB-NIR-RE | 0.987 | 0.536 | 0.762 | 0.0161 | 0.147 | 91.576 | 2.981 | 14.110 |
| 5 | WeedBananaCOD | RGB | 0.990 | 0.714 | 0.852 | 0.0268 | 0.108 | 91.720 | 3.128 | 22.423 |
| | | RGB-NIR | 0.990 | 0.719 | 0.855 | 0.0276 | 0.110 | 91.722 | 3.128 | 22.457 |
| | WeedMap | RGB | 0.986 | 0.527 | 0.756 | 0.0161 | 0.140 | 91.720 | 3.128 | 15.766 |
| | | RGB-NIR | 0.986 | 0.532 | 0.759 | 0.0165 | 0.149 | 91.722 | 3.128 | 15.790 |
| | | RGB-NIR-RE | 0.985 | 0.535 | 0.760 | 0.0167 | 0.146 | 91.724 | 3.128 | 15.813 |

Table 7. Comparison of the proposed FCB with other alternatives.

| Component | Dataset | Modality | $IoU_{BG}$ ↑ | $IoU_{W}$ ↑ | mIoU↑ | Inference time (s) | Training time (h) | Total params. (M) | Trainable params. (M) | GFLOPS |
|---|---|---|---|---|---|---|---|---|---|---|
| w/o FCB | WeedBananaCOD | RGB | 0.988 | 0.664 | 0.826 | 0.0223 | 0.093 | 89.871 | 1.280 | 14.053 |
| | | RGB-NIR | 0.989 | 0.693 | 0.841 | 0.0250 | 0.094 | 89.874 | 1.280 | 14.087 |
| | WeedMap | RGB | 0.978 | 0.514 | 0.746 | 0.0131 | 0.121 | 89.871 | 1.280 | 9.881 |
| | | RGB-NIR | 0.981 | 0.519 | 0.750 | 0.0134 | 0.125 | 89.874 | 1.280 | 9.905 |
| | | RGB-NIR-RE | 0.982 | 0.521 | 0.752 | 0.0134 | 0.130 | 89.876 | 1.280 | 9.928 |
| w/CBAM | WeedBananaCOD | RGB | 0.989 | 0.710 | 0.850 | 0.0382 | 0.140 | 90.048 | 1.457 | 14.056 |
| | | RGB-NIR | 0.988 | 0.693 | 0.840 | 0.0421 | 0.144 | 90.050 | 1.457 | 14.089 |
| | WeedMap | RGB | 0.978 | 0.521 | 0.750 | 0.0282 | 0.188 | 90.048 | 1.457 | 9.883 |
| | | RGB-NIR | 0.983 | 0.518 | 0.751 | 0.0286 | 0.194 | 90.050 | 1.457 | 9.906 |
| | | RGB-NIR-RE | 0.985 | 0.523 | 0.754 | 0.0292 | 0.202 | 90.052 | 1.457 | 9.930 |
| w/FCB | WeedBananaCOD | RGB | 0.991 | 0.733 | 0.862 | 0.0227 | 0.099 | 91.277 | 2.685 | 15.156 |
| | | RGB-NIR | 0.991 | 0.762 | 0.877 | 0.0298 | 0.105 | 91.279 | 2.685 | 15.190 |
| | WeedMap | RGB | 0.985 | 0.534 | 0.760 | 0.0141 | 0.128 | 91.277 | 2.685 | 10.657 |
| | | RGB-NIR | 0.986 | 0.545 | 0.766 | 0.0145 | 0.134 | 91.279 | 2.685 | 10.680 |
| | | RGB-NIR-RE | 0.987 | 0.547 | 0.767 | 0.0147 | 0.142 | 91.281 | 2.685 | 10.704 |

The performance of all FCBNet variants is presented in Table 8. The results show a consistent improvement in performance as the encoder scale increases, across both RGB and multispectral configurations in both datasets. In terms of efficiency, the expected increase in latency and training time is observed when scaling the architecture; however, the number of trainable parameters remains notably low due to the freezing strategy. In the Tiny variant, the number of trainable parameters is reduced from 30.6M to only 2.01M, representing a 93.4% reduction, while the Large variant achieves a massive reduction of 97.7%. This allows even the most complex model to be trained in approximately 0.2 hours, validating FCBNet as a flexible model whose structural efficiency enables the deployment of both lightweight and robust models under limited computational resources.

Table 8. Performance comparison of FCBNet across different ConvNeXt encoders.

| Encoder | Dataset | Modality | $IoU_{BG}$ ↑ | $IoU_{W}$ ↑ | mIoU↑ | Inference time (s) | Training time (h) | Total params. (M) | Trainable params. (M) | GFLOPS |
|---|---|---|---|---|---|---|---|---|---|---|
| ConvNeXt-tiny | WeedBananaCOD | RGB | 0.990 | 0.717 | 0.853 | 0.0139 | 0.060 | 30.604 | 2.015 | 13.151 |
| | | RGB-NIR | 0.991 | 0.740 | 0.865 | 0.0149 | 0.062 | 30.605 | 2.015 | 13.176 |
| | WeedMap | RGB | 0.985 | 0.537 | 0.761 | 0.0095 | 0.083 | 30.604 | 2.015 | 9.247 |
| | | RGB-NIR | 0.986 | 0.540 | 0.763 | 0.0103 | 0.091 | 30.605 | 2.015 | 9.264 |
| | | RGB-NIR-RE | 0.986 | 0.543 | 0.764 | 0.0107 | 0.098 | 30.607 | 2.015 | 9.282 |
| ConvNeXt-small | WeedBananaCOD | RGB | 0.987 | 0.658 | 0.822 | 0.0178 | 0.075 | 52.238 | 2.015 | 13.540 |
| | | RGB-NIR | 0.989 | 0.703 | 0.846 | 0.0200 | 0.080 | 52.240 | 2.015 | 13.565 |
| | WeedMap | RGB | 0.985 | 0.540 | 0.762 | 0.0132 | 0.107 | 52.238 | 2.015 | 9.520 |
| | | RGB-NIR | 0.986 | 0.544 | 0.765 | 0.0144 | 0.113 | 52.240 | 2.015 | 9.538 |
| | | RGB-NIR-RE | 0.987 | 0.545 | 0.766 | 0.0133 | 0.118 | 52.241 | 2.015 | 9.556 |
| ConvNeXt-base | WeedBananaCOD | RGB | 0.991 | 0.733 | 0.862 | 0.0227 | 0.099 | 91.277 | 2.685 | 15.156 |
| | | RGB-NIR | 0.991 | 0.762 | 0.877 | 0.0298 | 0.105 | 91.279 | 2.685 | 15.190 |
| | WeedMap | RGB | 0.985 | 0.534 | 0.760 | 0.0141 | 0.128 | 91.277 | 2.685 | 10.657 |
| | | RGB-NIR | 0.986 | 0.545 | 0.766 | 0.0145 | 0.134 | 91.279 | 2.685 | 10.680 |
| | | RGB-NIR-RE | 0.987 | 0.547 | 0.767 | 0.0147 | 0.142 | 91.281 | 2.685 | 10.704 |
| ConvNeXt-large | WeedBananaCOD | RGB | 0.991 | 0.746 | 0.868 | 0.0360 | 0.169 | 202.322 | 4.555 | 19.504 |
| | | RGB-NIR | 0.992 | 0.771 | 0.881 | 0.0381 | 0.171 | 202.325 | 4.555 | 19.554 |
| | WeedMap | RGB | 0.987 | 0.546 | 0.766 | 0.0252 | 0.201 | 202.322 | 4.555 | 13.714 |
| | | RGB-NIR | 0.986 | 0.548 | 0.767 | 0.0251 | 0.211 | 202.325 | 4.555 | 13.749 |
| | | RGB-NIR-RE | 0.987 | 0.551 | 0.769 | 0.0257 | 0.215 | 202.329 | 4.555 | 13.784 |

Finally, Table 9 presents the comparison between the proposed architecture and several state of the art methods. The results show that FCBNet-large achieves the highest performance across both datasets and all spectral modalities, suggesting a superior ability to adapt and effectively leverage information from multiple channels. It is also notable that, in the WeedMap dataset, all evaluated models exhibit a lower performance ceiling compared to WeedBananaCOD; however, FCBNet consistently outperforms its competitors even under these data constraints.

On the other hand, although the FCBNet-tiny variant is slightly outperformed in mIoU by models such as DeepLabV3+ or SegFormer in certain scenarios, such as the WeedBananaCOD dataset, the performance gap is minimal compared to the substantial efficiency gains. While these models require training times exceeding 0.12 and 0.31 hours respectively and exhibit significantly higher GFLOPs, FCBNet-tiny achieves competitive performance with only 0.06 hours of training. Overall, these results validate the superiority of the proposed approach, establishing it as a highly efficient solution that drastically reduces computational requirements without compromising accuracy compared to the other heavy methods in the literature.

To complement the quantitative analysis, Figs. 5a and 5b present a visual comparison of inference on the WeedBananaCOD and WeedMap datasets, respectively, using their full modalities. In WeedBananaCOD, FCBNet achieves more precise instance delineation and a significant reduction in false detections (rows b and d), producing results that more closely match the ground truth compared to the other models. In WeedMap, despite the sparse distribution and small size of weed instances, the proposed architecture also demonstrates a greater ability to identify small objects with a lower error rate. In both cases, the qualitative results confirm that the FCBNet correction mechanism better preserves fine details and improves the spatial coherence of camouflaged weed segmentation.

## 5. Limitations and Future work

Despite our extensive evaluation, we acknowledge some limitations. First, the evaluation can be extended to other datasets containing additional scenarios, spectral bands, and classes (crops or different weed species), as this study focused on binary segmentation. Also, the comparative analysis should be expanded to include methods beyond those that were currently feasible for reimplementation.

Furthermore, evaluating FCBNet on specialized hardware and embedded systems, such as Raspberry Pi or UAV platforms, would provide valuable complementary validation of its behaviour under real deployment conditions. Finally, exploring structural modifications, including dual encoders for multispectral processing or alternative decoder designs, represents a natural direction for future research.

Table 9. FCBNet compared to other approaches from the literature. We have selected the best and lightest FCBNet model for comparison.

| Encoder | Dataset | Modality | $IoU_{BG}\uparrow$ | $IoU_{W}\uparrow$ | mIoU↑ | Inference time (s) | Training time (h) | Total params. (M) | Trainable params. (M) | GFLOPS |
|---|---|---|---|---|---|---|---|---|---|---|
| FCBNet-tiny | WeedBananaCOD | RGB | 0.990 | 0.717 | 0.853 | 0.0139 | 0.060 | 30.604 | 2.015 | 13.151 |
| | | RGB-NIR | 0.991 | 0.740 | 0.865 | 0.0149 | 0.062 | 30.605 | 2.015 | 13.176 |
| | WeedMap | RGB | 0.985 | 0.537 | 0.761 | 0.0095 | 0.083 | 30.604 | 2.015 | 9.247 |
| | | RGB-NIR | 0.986 | 0.540 | 0.763 | 0.0103 | 0.091 | 30.605 | 2.015 | 9.264 |
| | | RGB-NIR-RE | 0.986 | 0.543 | 0.764 | 0.0107 | 0.098 | 30.607 | 2.015 | 9.282 |
| FCBNet-large | WeedBananaCOD | RGB | 0.991 | 0.746 | 0.868 | 0.0360 | 0.169 | 202.322 | 4.555 | 19.504 |
| | | RGB-NIR | 0.992 | 0.771 | 0.881 | 0.0381 | 0.171 | 202.325 | 4.555 | 19.554 |
| | WeedMap | RGB | 0.987 | 0.546 | 0.766 | 0.0252 | 0.201 | 202.322 | 4.555 | 13.714 |
| | | RGB-NIR | 0.986 | 0.548 | 0.767 | 0.0251 | 0.211 | 202.325 | 4.555 | 13.749 |
| | | RGB-NIR-RE | 0.987 | 0.551 | 0.769 | 0.0257 | 0.215 | 202.329 | 4.555 | 13.784 |
| WeedSense [33] | WeedBananaCOD | RGB | 0.984 | 0.596 | 0.790 | 0.0223 | 0.089 | 12.721 | 12.721 | 12.731 |
| | | RGB-NIR | 0.986 | 0.625 | 0.805 | 0.0225 | 0.093 | 12.723 | 12.723 | 12.778 |
| | WeedMap | RGB | 0.979 | 0.471 | 0.725 | 0.0115 | 0.138 | 12.721 | 12.721 | 8.952 |
| | | RGB-NIR | 0.984 | 0.515 | 0.749 | 0.0190 | 0.151 | 12.723 | 12.723 | 8.985 |
| | | RGB-NIR-RE | 0.984 | 0.519 | 0.752 | 0.0196 | 0.166 | 12.724 | 12.724 | 9.018 |
| DeepLabV3+ [3] | WeedBananaCOD | RGB | 0.988 | 0.721 | 0.855 | 0.0270 | 0.123 | 26.678 | 26.678 | 36.762 |
| | | RGB-NIR | 0.991 | 0.725 | 0.858 | 0.0390 | 0.125 | 26.681 | 26.681 | 36.968 |
| | WeedMap | RGB | 0.983 | 0.509 | 0.746 | 0.0190 | 0.161 | 26.678 | 26.678 | 25.849 |
| | | RGB-NIR | 0.984 | 0.515 | 0.749 | 0.0190 | 0.170 | 26.681 | 26.681 | 25.993 |
| | | RGB-NIR-RE | 0.986 | 0.534 | 0.760 | 0.0200 | 0.189 | 26.684 | 26.684 | 26.137 |
| U-Net [29] | WeedBananaCOD | RGB | 0.990 | 0.715 | 0.851 | 0.0243 | 0.097 | 32.521 | 32.521 | 42.771 |
| | | RGB-NIR | 0.990 | 0.720 | 0.855 | 0.0257 | 0.100 | 32.524 | 32.524 | 42.976 |
| | WeedMap | RGB | 0.975 | 0.427 | 0.701 | 0.0109 | 0.119 | 32.521 | 32.521 | 30.073 |
| | | RGB-NIR | 0.981 | 0.508 | 0.745 | 0.0112 | 0.126 | 32.524 | 32.524 | 30.218 |
| | | RGB-NIR-RE | 0.986 | 0.550 | 0.768 | 0.0117 | 0.131 | 32.528 | 32.528 | 30.362 |
| SegFormer [41] | WeedBananaCOD | RGB | 0.989 | 0.695 | 0.842 | 0.0203 | 0.310 | 84.595 | 84.595 | 49.733 |
| | | RGB-NIR | 0.990 | 0.738 | 0.864 | 0.0295 | 0.312 | 84.598 | 84.598 | 49.784 |
| | WeedMap | RGB | 0.986 | 0.506 | 0.746 | 0.0177 | 0.420 | 84.595 | 84.595 | 34.116 |
| | | RGB-NIR | 0.985 | 0.532 | 0.758 | 0.0180 | 0.421 | 84.598 | 84.598 | 34.153 |
| | | RGB-NIR-RE | 0.985 | 0.535 | 0.760 | 0.0189 | 0.421 | 84.601 | 84.601 | 34.189 |
| SK-U-Net [25] | WeedBananaCOD | RGB | 0.986 | 0.641 | 0.813 | 0.0231 | 0.170 | 34.444 | 34.444 | 44.745 |
| | | RGB-NIR | 0.990 | 0.705 | 0.847 | 0.0238 | 0.170 | 34.447 | 34.447 | 44.950 |
| | WeedMap | RGB | 0.983 | 0.532 | 0.758 | 0.0140 | 0.160 | 34.444 | 34.444 | 31.461 |
| | | RGB-NIR | 0.982 | 0.539 | 0.761 | 0.0164 | 0.170 | 34.447 | 34.447 | 31.606 |
| | | RGB-NIR-RE | 0.984 | 0.539 | 0.762 | 0.0140 | 0.178 | 34.450 | 34.450 | 31.751 |

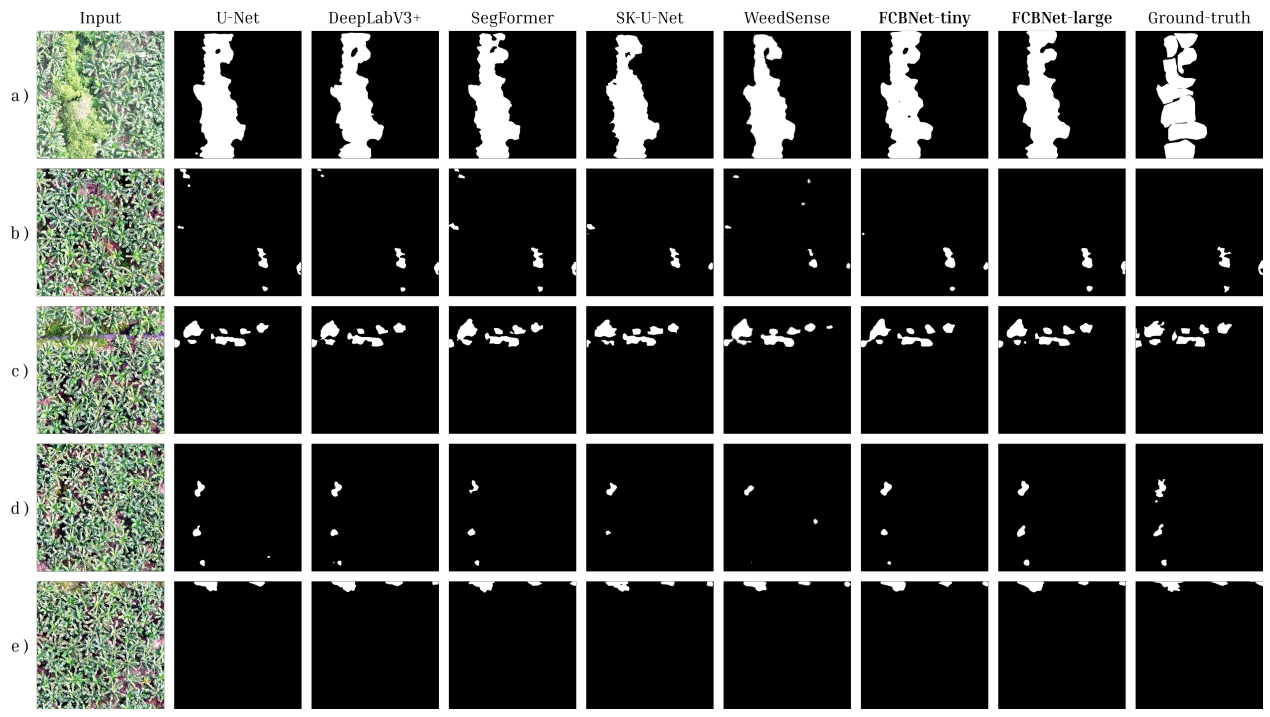

(a) WeedBananaCOD (RGB-NIR).

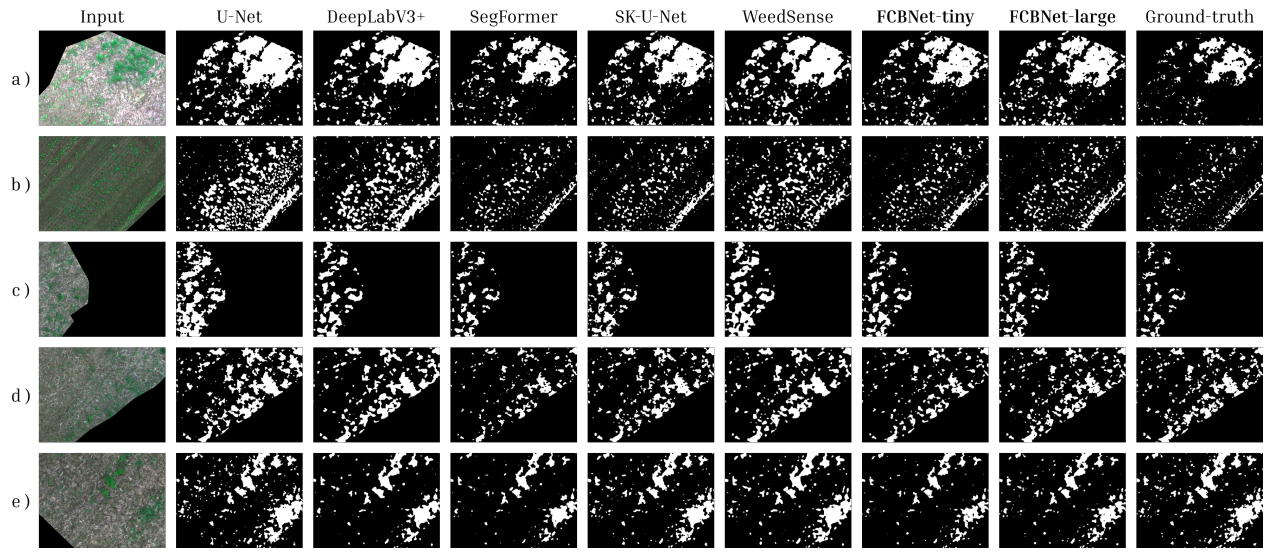

(b) WeedMap (RGB-NIR-RE).

Figure 5. Qualitative comparison between FCBNet and other models from the state-of-the-art. Zoom in for better visualization.

# 6. Conclusions

We introduce FCBNet, an efficient architecture for weed detection. It employs a frozen ConvNeXt backbone integrated with the proposed FCB block, which combines efficient convolutions to refine feature representations at minimal computational cost, along with a lightweight FPN decoder. Extensive evaluations on the WeedBananaCOD and WeedMap datasets demonstrate that FCBNet outperforms established models such as U-Net, SK-U-Net, SegFormer, and WeedSense in both mIoU and operational efficiency. Specifically, the freezing strategy enables a reduction of trainable parameters >93%, achieving training times of only 0.06 to 0.2 hours without compromising performance.

# Acknowledgement

This material is based upon work supported by the Air Force Office of Scientific Research under award number FA9550-24-1-0206; partially supported by the Grant PID2021-128945NB-I00 funded by MICIU/AEI/10.13039/501100011033 and by ERDF/EU; and by Grant PID2024-162815NB-I00 funded by MICIU/AEI/10.13039/501100011033 and by ERDF/EU. This work is also supported by the Generalitat de Catalunya under grant SDC007/25/000005 co-funded by the European Union - NextGenerationEU, and by the CERCA Program supporting CVC's general activities.